\documentclass[twoside,11pt]{article}

\usepackage{subfigure}
\usepackage{array}
\usepackage{booktabs}
\usepackage{enumitem}
\usepackage{graphicx}
\usepackage{algorithm}
\usepackage{algorithmic}
\usepackage{amsthm}
\usepackage{multirow}
\newtheorem{assumption}{Assumption}
\usepackage[abbrvbib, preprint]{ArXiv}
\usepackage[svgnames]{xcolor}
\hypersetup{
hidelinks,
colorlinks=true,
linkcolor=Indigo,
urlcolor=DeepSkyBlue4,
citecolor=Indigo
}

\usepackage{amsmath}

\DeclareFixedFont{\myfont}{OT1}{ptm}{m}{n}{8pt}
\DeclareFixedFont{\myfontb}{OT1}{ptm}{bx}{n}{8pt}

\usepackage{marvosym}
\makeatletter
\def\@fnsymbol#1{\ifcase#1\or \text{\Letter}\or *\or \dagger\or \ddagger\else\@arabic{#1}\fi}
\makeatother

\usepackage{lastpage}
\jmlrheading{23}{\ Preprint 2026}{1-\pageref{LastPage}}{1/21; Revised 5/22}{9/22}{21-0000}{}

\firstpageno{1}

\begin{document}

\title{Breaking the Prototype Bias Loop: Confidence-Aware Federated Contrastive Learning for Highly Imbalanced Clients}

\author{\name Tian-Shuang Wu \email tianshuangwu@hhu.edu.cn \\
       \addr Key Laboratory of Water Big Data Technology of Ministry of Water Resources,\\ College of Computer Science and Software Engineering, Hohai University, Nanjing, China
       \AND
       \name Shen-Huan Lyu~\thanks{Corresponding author} \email lvsh@hhu.edu.cn \\
       \addr Key Laboratory of Water Big Data Technology of Ministry of Water Resources,\\ College of Computer Science and Software Engineering, Hohai University, Nanjing, China\\
       \addr Department of Computer Science, City University of Hong Kong, Hong Kong, China\\
       \addr State Key Laboratory for Novel Software Technology,
       Nanjing University, Nanjing, China
       \AND
       \name Ning Chen \email che-n-ing@hhu.edu.cn \\
       \addr Key Laboratory of Water Big Data Technology of Ministry of Water Resources,\\ College of Computer Science and Software Engineering, Hohai University, Nanjing, China
       \AND
       \name Yi-Xiao He \email heyx@njucm.edu.cn \\
       \addr School of Artificial Intelligence and Information Technology, Nanjing University of Chinese Medicine, Nanjing, China
       \AND
       \name Bin Tang \email cstb@hhu.edu.cn \\       
       \addr Key Laboratory of Water Big Data Technology of Ministry of Water Resources,\\ College of Computer Science and Software Engineering, Hohai University, Nanjing, China
       \AND
       \name Baoliu Ye \email yebl@nju.edu.cn \\
       \addr State Key Laboratory for Novel Software Technology,
       Nanjing University, Nanjing, China
       \AND
       \name Qingfu Zhang \email qingfu.zhang@cityu.edu.hk\\
       \addr Department of Computer Science, City University of Hong Kong, Hong Kong, China\\
       }
\editor{My editor}

\maketitle

\begin{abstract}
Local class imbalance and data heterogeneity across clients often trap prototype-based federated contrastive learning in a prototype bias loop: biased local prototypes induced by imbalanced data are aggregated into biased global prototypes, which are repeatedly reused as contrastive anchors, accumulating errors across communication rounds. To break this loop, we propose Confidence-Aware Federated Contrastive Learning (CAFedCL), a novel framework that improves the prototype aggregation mechanism and strengthens the contrastive alignment guided by prototypes. CAFedCL employs a confidence-aware aggregation mechanism that leverages predictive uncertainty to downweight high-variance local prototypes. In addition, generative augmentation for minority classes and geometric consistency regularization are integrated to stabilize the structure between classes. From a theoretical perspective, we provide an expectation-based analysis showing that our aggregation reduces estimation variance, thereby bounding global prototype drift and ensuring convergence. Extensive experiments under varying levels of class imbalance and data heterogeneity demonstrate that CAFedCL consistently outperforms representative federated baselines in both accuracy and client fairness.
\end{abstract}


\begin{figure}[t]
\centering
\includegraphics[width=\linewidth]{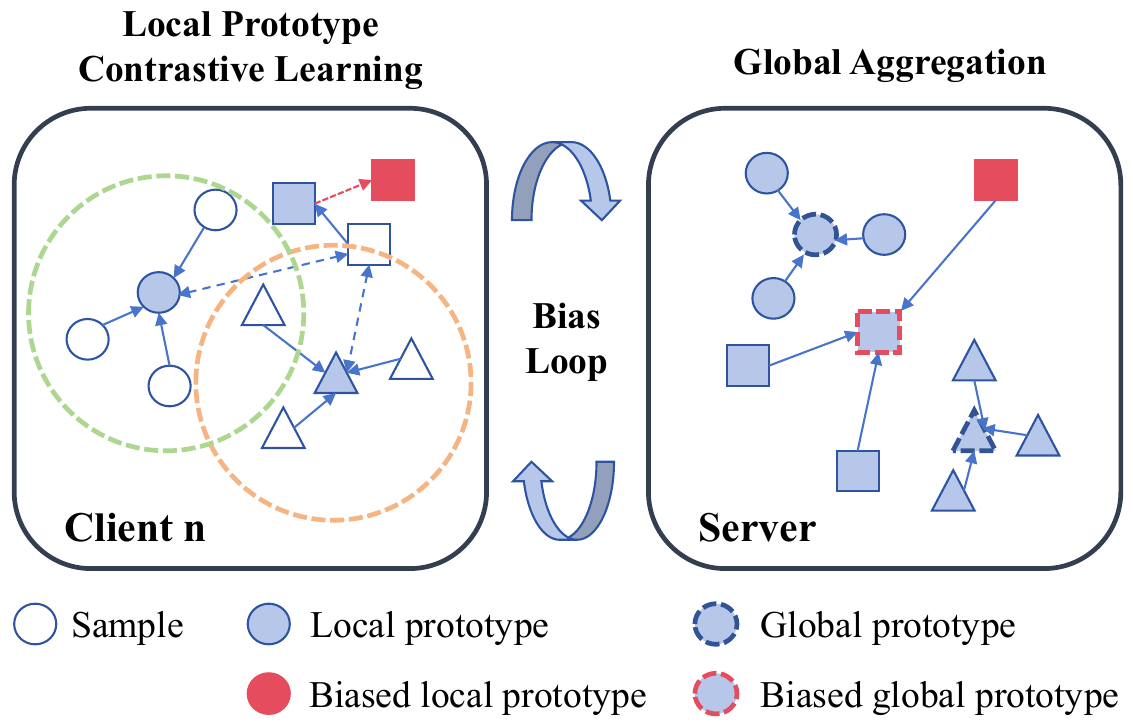}
\caption{The bias loop in federated contrastive learning.}
\label{fig:introduction}
\end{figure}

\section{Introduction}

Federated Learning (FL)~\citep{mcmahan2017communication} enables collaborative model training without centralizing raw data, and has become a common paradigm for privacy-preserving analytics. 
In practice, however, vanilla FL is often challenged by statistical heterogeneity, especially label distribution skew and extreme class imbalance (long-tailed data)~\citep{imteaj2021survey, tan2022towards}. 
Recent studies~\citep{huang2025fedcosr} further suggest that label heterogeneity frequently co-occurs with data scarcity for minority classes, resulting in degraded generalization and uneven performance across clients.
While personalized FL has made progress, many approaches still focus on aggregating generic model parameters, and may not fully exploit label-structured information that is critical under skewed local supervision~\citep{kairouz2021advances}.

To alleviate client drift, a growing line of work shifts from parameter-space aggregation to representation-space alignment.
Federated contrastive learning (FedCL) is a representative direction that regularizes local embeddings toward more consistent global representations.
However, standard contrastive learning typically relies on large sets of negative pairs or large batch sizes, which are difficult to maintain in privacy-preserving FL.
This has motivated prototype-based federated contrastive learning, where prototypes are class-wise feature centroids that serve as lightweight summaries of class semantics~\citep{mu2023fedproc, tan2022fedproto}.
By communicating global prototypes as shared semantic anchors, clients can align their local representations with global class structure at low communication cost, without transmitting individual samples~\citep{tan2022federated}.
Beyond single-prototype schemes, multi-prototype variants further model intra-class diversity while still enabling inter-client knowledge transfer~\citep{qiao2023mp}.

Despite its appeal, this paradigm implicitly assumes that aggregated global prototypes reliably approximate true class centers.
We argue that this assumption can be violated under extreme imbalance and heterogeneous client label quality, which commonly arises in applications such as medical screening (rare positives) and industrial defect inspection (rare faults).
In such regimes, prototype errors may become self-reinforcing across rounds: biased or high-variance local prototypes contaminate the aggregated global anchors, and the contaminated anchors are then reused to guide subsequent contrastive updates. 
This feedback loop can progressively distort the prototype geometry and harm minority-class discrimination. 
We refer to this phenomenon as the \textit{Prototype Bias Loop} (Fig.~\ref{fig:introduction}).

To address this issue, we propose \textbf{CAFedCL}, a confidence-aware federated prototypical contrastive learning framework tailored to extreme imbalance and heterogeneous client quality.
The key idea is to treat prototypes as uncertain estimates rather than deterministic targets.
On each client, we optionally augment minority classes using a conditional GAN to mitigate severe data scarcity, and train the encoder with a prototype-guided contrastive objective complemented by a global-guided geometric regularizer that preserves inter-class separation.
On the server, we replace naive averaging with confidence-weighted aggregation for both prototypes and encoder parameters.
Each client reports class-wise confidence scores that combine generation quality signals with validation-based reliability, so that unreliable client--class contributions are down-weighted during aggregation, leading to more stable global anchors.

The main contributions of this work are summarized as follows:
\begin{itemize}
    \item We identify a failure mechanism of prototype-based federated contrastive learning under extreme imbalance, showing how cross-round anchor reuse and naive averaging can amplify prototype errors, forming the Prototype Bias Loop.
    \item We propose CAFedCL, which stabilizes minority representations via optional tail augmentation and geometric regularization, and mitigates unreliable updates via class-wise confidence-weighted aggregation for prototypes and model parameters.
    \item We conduct extensive experiments across diverse non-IID and long-tailed settings, demonstrating consistent improvements in accuracy and client fairness over strong baselines with modest overhead.
\end{itemize}

\section{Related Work}

\subsection{Federated learning}
Federated learning (FL) has emerged as a privacy-preserving distributed learning paradigm that enables multiple clients to collaboratively optimize a shared model without transferring raw data to a centralized server~\citep{mcmahan2017communication}. This decentralized training protocol has become increasingly important in data-sensitive applications, such as medical images~\citep{adnan2022federated, liu2021feddg} and Internet-of-Things (IoT) systems~\citep{wang2025lightweight, khan2021federated}, where data are naturally generated and stored across geographically dispersed devices under resource constraints~\citep{imteaj2021survey, kairouz2021advances}.

Despite its promise, real-world FL deployments are often hindered by severe system and statistical heterogeneity~\citep{luo2021no}, particularly the non-IID nature of client data, which can cause client drift, objective inconsistency, and degraded generalization compared to centralized training \citep{zhao2018federated,li2020federated}. To address these challenges, substantial progress has been made from both the optimization and personalization perspectives. Representative solutions include regularizing local updates to stabilize training~\citep{li2020federated}, correcting client drift with control variates~\citep{karimireddy2020scaffold}, and mitigating objective inconsistency via normalized aggregation~\citep{wang2020tackling, diao2024exploiting, li2021fedbn}. Moreover, recent surveys have summarized the advances and highlighted persistent open problems, motivating the community to further develop robust and communication-efficient federated algorithms for heterogeneous environments~\citep{kairouz2021advances,tan2022towards}.

\subsection{Contrastive learning in FL}
Contrastive learning is a powerful paradigm for learning transferable representations by pulling semantically similar pairs together while separating negatives, and has achieved remarkable success in both self-supervised and supervised settings \citep{oord2018representation, chen2020simple, he2020momentum}. Motivated by its generalization benefits, recent works introduce contrastive objectives into federated learning to mitigate representation misalignment under non-IID data and improve robustness against client drift \citep{zhang2021federated, guo2023fedbr, zhuang2022divergence}. Compared with optimizing task-specific losses alone, federated contrastive learning (FedCL) provides an additional representation-level regularization and has become a promising direction for heterogeneous FL.

Existing FedCL approaches mainly differ in how contrastive signals are constructed and shared under communication and privacy constraints. Tan et al. explore contrastive regularization for federated learning from pre-trained models \citep{tan2022federated}, while model-contrastive federated learning directly contrasts client models to promote global consistency \citep{li2021model}. Prototype-centric methods further use class-level representations as shared anchors \citep{dai2023tackling}, including FedProto for prototype aggregation and broadcasting \citep{tan2022fedproto} and FedProc with prototypical contrastive objectives \citep{mu2023fedproc}. Beyond single-prototype modeling, Mp-FedCL employs multi-prototypes to capture intra-class diversity \citep{qiao2023mp}, and FedTGP introduces trainable global prototypes with adaptive-margin contrastive learning to address both data and model heterogeneity \citep{zhang2024fedtgp}. Moreover, FedSC provides a provable federated self-supervised framework based on spectral contrastive objectives with differential privacy protection \citep{jing2024fedsc}, and FedRCL adopts a relaxed contrastive loss with an over-similarity penalty to prevent collapse under heterogeneity \citep{seo2024relaxed}. These efforts highlight the effectiveness of contrastive learning for federated representation alignment.

Despite these advances, FedCL remains challenging in class-imbalanced and label-skewed settings, where minority classes are scarce and client label distributions are highly uneven. In such regimes, naive contrastive learning and prototype aggregation can be dominated by majority features, resulting in biased prototypes \citep{li2021adaptive}, degraded minority discrimination, and even error amplification through iterative global aggregation \citep{wang2021addressing, wu2023fediic, li2021fedbn}. Developing imbalance-aware FedCL mechanisms that preserve robust minority representations and stabilize global coordination therefore remains an important yet under-explored direction.

\section{Preliminaries}
\subsection{Federated Prototype Learning}
We consider a federated system with $K$ clients indexed by $k\in\{1,\dots,K\}$.
Client $k$ holds a private dataset $\mathcal{D}_k=\{(x_i,y_i)\}_{i=1}^{N_k}$ with labels $y_i\in\{1,\dots,C\}$.
Let $f(\cdot;\theta_k)$ denote the client encoder.
Federated Prototype Learning (FPL) exchanges lightweight class-wise representatives (\emph{prototypes}) instead of full model information.
For class $c$, the local prototype on client $k$ is computed as
\begin{equation}
\mathbf{p}_{k,c}=\frac{1}{|\mathcal{D}_{k,c}|}\sum_{(x,y)\in\mathcal{D}_{k,c}} f(x;\theta_k),
\label{eq:local_proto}
\end{equation}
where $\mathcal{D}_{k,c}\subseteq\mathcal{D}_k$ contains samples of class $c$ and $|\mathcal{D}_{k,c}|=:n_{k,c}$.
The server aggregates participating client prototypes to obtain global prototypes $\mathcal{P}_g=\{\mathbf{p}_{g,c}\}_{c=1}^C$:
\begin{equation}
\mathbf{p}_{g,c}=\frac{\sum_{k\in S_t} n_{k,c}\mathbf{p}_{k,c}}{\sum_{k\in S_t} n_{k,c}},
\label{eq:global_agg}
\end{equation}
where $S_t$ is the set of participating clients at round $t$.
When prototypes are used for cosine similarity, we apply $\ell_2$ normalization to embeddings and prototypes.

\subsection{Estimation Bias under Imbalance}
In realistic FL, client data are often non-IID and long-tailed.
For minority classes, the sample size $n_{k,c}$ can be extremely small, making the empirical prototype in Eq.~(\ref{eq:local_proto}) a high-variance estimator of the underlying class center $\boldsymbol{\mu}_{k,c}$.

Assuming bounded intra-class feature variance, the prototype estimation error typically scales as
\begin{equation}
\mathbb{E}\big[\|\mathbf{p}_{k,c}-\boldsymbol{\mu}_{k,c}\|_2^2\big]\ \lesssim\ \frac{\sigma_{k,c}^2}{n_{k,c}}.
\label{eq:drift_bound}
\end{equation}
Thus, tail client--class pairs produce unreliable prototypes.
Naive aggregation in Eq.~(\ref{eq:global_agg}) does not explicitly account for this reliability gap, allowing noisy local prototypes to contaminate global anchors, which will be repeatedly reused for supervision in prototype-guided contrastive learning.

\section{Method}
In this section, we propose CAFedCL, a prototype-based federated contrastive framework tailored for extreme label imbalance and heterogeneous clients.
Our design is motivated by a key observation: under long-tailed clients, naive class-wise prototype aggregation may inject high-variance and biased anchors into the server; once broadcast, these anchors are repeatedly reused in prototype-guided contrastive learning, forming a self-reinforcing prototype bias loop. To break this loop, we introduce a class-wise confidence mechanism that down-weights unreliable client classes during both prototype and model aggregation, while two lightweight stabilizers improve minority representations.

\subsection{Prototype-guided Contrastive Learning for Federated Alignment}
\label{sec:method_contrastive}

Federated learning under non-IID and long-tailed settings often suffers from representation misalignment, where local encoders drift toward client-specific features.
Prototype-guided federated contrastive learning addresses this issue by using global prototypes as lightweight semantic anchors: clients align their embeddings to shared class references without exchanging raw data or full model logits.

Formally, for a sample $(x_i,y_i)$ on client $k$, we denote its normalized embedding as $\mathbf{z}_i=f(x_i;\theta_k)$ (followed by $\ell_2$ normalization).
The server maintains global prototypes $\mathcal{P}_g=\{\mathbf{p}_{g,c}\}_{c=1}^C$, where each $\mathbf{p}_{g,c}$ represents the global class-$c$ center in the embedding space.
Compared with purely local supervised contrastive learning, prototype guidance is particularly effective when local batches are small or classes are missing on some clients.

A typical prototype-guided contrastive objective, denoted as $\mathcal{L}_{\mathrm{I2P}}$ (instance-to-prototype), aligns each embedding to its class prototype while repelling other class prototypes:
\begin{equation}
    \mathcal{L}_{\mathrm{I2P}} = - \frac{1}{|\mathcal{B}|} \sum_{(x_i, y_i) \in \mathcal{B}} 
    \log \frac{\exp(\text{sim}(\mathbf{z}_i, \mathbf{p}_{g, y_i}) / \tau)}
    {\sum_{c=1}^C \exp(\text{sim}(\mathbf{z}_i, \mathbf{p}_{g, c}) / \tau)},
    \label{eq:i2p_loss}
\end{equation}
where $\mathcal{B}$ is the local mini-batch, $\text{sim}(\cdot,\cdot)$ denotes cosine similarity, and $\tau$ is a temperature.
Minimizing Eq.~(\ref{eq:i2p_loss}) pulls client embeddings toward unified global anchors, thereby mitigating cross-client representation drift.

After local training, each client computes class-wise prototypes $\mathbf{p}_{k,c}^{t+1}$ using Eq.~(\ref{eq:local_proto}).
When used in contrastive similarity, we apply $\ell_2$ normalization to both embeddings and prototypes, i.e., $\tilde{\mathbf{z}}=\mathrm{norm}(f(x;\theta_k))$ and $\tilde{\mathbf{p}}_{k,c}=\mathrm{norm}(\mathbf{p}_{k,c})$, and the server broadcasts normalized global prototypes $\tilde{\mathbf{p}}_{g,c}$.
This keeps the contrastive objective consistent with cosine similarity in Eq.~(\ref{eq:i2p_loss}).

\subsection{Prototype Bias Loop in Federated Contrastive Learning}
\label{sec:bias_loop}

Section~\ref{sec:method_contrastive} shows that broadcasting global prototypes provides lightweight semantic anchors for instance--prototype contrast.
However, under extreme label imbalance and heterogeneous client quality, local prototypes can be noisy or biased.
Once aggregated, these imperfect prototypes become global anchors and are repeatedly reused in Eq.~(\ref{eq:i2p_loss}), so the error may persist and accumulate over rounds, forming a self-reinforcing prototype bias loop.

To make this explicit, we fix a class $c$ and track the mean squared error of the global prototype $\mathbf{p}^{t}_{g,c}$ with respect to the underlying class center $\boldsymbol{\mu}_c$:
\(
E^t \triangleq \mathbb{E}\|\mathbf{p}^{t}_{g,c}-\boldsymbol{\mu}_c\|^2.
\)

\begin{assumption}[Bounded intra-class variance]
\label{ass:variance}
For any class $c$ on client $k$, embeddings have bounded variance around a client-specific center $\boldsymbol{\mu}_{k,c}$:
\(
\mathbb{E}_{x\sim \mathbf{p}_{k,c}} \| f_{\theta_k}(x) - \boldsymbol{\mu}_{k,c} \|^2 \le \sigma^2_{k,c}.
\)
\end{assumption}

Under Assumption~\ref{ass:variance}, the quality of a local prototype is fundamentally constrained by the effective sample size. Recall the local prototype definition in Eq.~(\ref{eq:local_proto}). We have the following estimation bound.

\begin{lemma}
\label{lem:local_uncertainty}
Under Assumption~\ref{ass:variance}, the estimation error of the local prototype satisfies
\begin{equation}
\mathbb{E}\|\mathbf{p}_{k,c}-\boldsymbol{\mu}_{k,c}\|^2 \le \frac{\sigma^2_{k,c}}{n^{\mathrm{eff}}_{k,c}},
\end{equation}
where $n^{\mathrm{eff}}_{k,c}=n_{k,c}+\gamma m_{k,c}$ is the effective sample size, $m_{k,c}$ is the number of synthetic samples (if any), and $\gamma\in[0,1]$ discounts their reliability.
\end{lemma}

Lemma~\ref{lem:local_uncertainty} implies that tail client--class pairs with small $n^{\mathrm{eff}}_{k,c}$ produce high-variance prototypes.
Consequently, naive aggregation can be fragile: even a few unreliable prototypes may inject noticeable noise into the global anchor.

The effect is further amplified by \emph{anchor reuse}.
In prototype-guided FedCL, the global prototype $\mathbf{p}^{t}_{g,c}$ is broadcast and repeatedly reused in Eq.~(\ref{eq:i2p_loss}), which continuously pulls local representations toward the current anchor.
As a first-order abstraction, we model this by a contractive dynamics:
\begin{equation}
\mathbf{p}^{t+1}_{k,c} = (1-\rho)\tilde{\boldsymbol{\mu}}_{k,c} + \rho\,\mathbf{p}^{t}_{g,c} + \boldsymbol{\xi}^{t+1}_{k,c},
\end{equation}
where $\tilde{\boldsymbol{\mu}}_{k,c}$ denotes the class-$c$ center induced by local data and optimization \emph{without} reusing the previous global anchor, 
$\rho\in[0,1)$ summarizes the anchor attraction strength introduced by prototype reuse in Eq.~(\ref{eq:i2p_loss}), 
and $\boldsymbol{\xi}^{t+1}_{k,c}$ is zero-mean noise with $\mathbb{E}\|\boldsymbol{\xi}^{t+1}_{k,c}\|^2 \le \sigma^2_{k,c}/n^{\mathrm{eff}}_{k,c}$.

On the server, the global prototype is obtained by class-wise weighted aggregation over participating clients $S_t$:
\begin{equation}
\mathbf{p}^{t+1}_{g,c} = \sum_{k\in S_t}\alpha^t_{k,c}\, \mathbf{p}^{t+1}_{k,c},
\quad \alpha^t_{k,c}\ge 0,\ \ \sum_{k\in S_t}\alpha^t_{k,c}=1.
\end{equation}

\begin{proposition}
\label{prop:bias_loop}
Let $E^t=\mathbb{E}\|\mathbf{p}^{t}_{g,c}-\boldsymbol{\mu}_c\|^2$.
Under Assumption~\ref{ass:variance} and the dynamics above, we have
\begin{equation}
\begin{aligned}
&E^{t+1}
&\le\ 
\underbrace{\rho^2 E^t}_{\text{anchor feedback}}
+
\underbrace{2(1-\rho)^2\,\Delta^t_{\mathrm{bias}}}_{\text{heterogeneity gap}}
+
\underbrace{2\sum_{k\in S_t}(\alpha^t_{k,c})^2\frac{\sigma^2_{k,c}}{n^{\mathrm{eff}}_{k,c}}}_{\text{variance injection}},
\end{aligned}
\label{eq:bias_rec}
\end{equation}
where
\(
\Delta^t_{\mathrm{bias}}
=
\big\|
\sum_{k\in S_t}\alpha^t_{k,c}(\tilde{\boldsymbol{\mu}}_{k,c}-\boldsymbol{\mu}_c)
\big\|^2
\)
measures the non-IID shift of class centers across clients.
\end{proposition}

\begin{remark}
Proposition~\ref{prop:bias_loop} decomposes the global prototype error into three sources.
\emph{Anchor feedback} ($\rho^2 E^t$) shows that once a biased global prototype appears, it is propagated to subsequent rounds because $\mathbf{p}^t_{g,c}$ is reused as the contrastive anchor in Eq.~(\ref{eq:i2p_loss}).
\emph{Heterogeneity gap} ($\Delta^t_{\mathrm{bias}}$) captures the unavoidable center shift induced by non-IID local training.
Crucially, the \emph{variance injection} term scales with $(\alpha^t_{k,c})^2/n^{\mathrm{eff}}_{k,c}$ and is amplified by tail client--class pairs with small $n^{\mathrm{eff}}_{k,c}$, making naive aggregation fragile under long-tailed data.
CAFedCL targets this dominant term by replacing $\alpha^t_{k,c}$ with class-wise confidence weights $\mathrm{conf}^t_{k,c}$ in prototype aggregation, so that unreliable client--class prototypes contribute less to the global anchor used in the next-round contrastive training.
\end{remark}

All proofs are in Appendix. A simple observation is that reliability-aware weighting (e.g., proportional to $n^{\mathrm{eff}}_{k,c}$) reduces the effective squared weights in the variance-injection term. We provide a formal statement in Appendix.

\subsection{Class-wise Confidence-weighted Aggregation}
\label{sec:conf_agg}

Proposition~\ref{prop:bias_loop} indicates that the prototype bias loop is mainly amplified by the variance-injection term, dominated by unreliable client--class pairs under long-tailed data.
To suppress this effect, CAFedCL assigns each client prototype a \emph{class-wise confidence} score and uses it as a reliability weight during aggregation.
Different from purely frequency-based weighting, our confidence combines multiple complementary signals so that a client can be down-weighted even when its sample count is not extremely small but its prototype quality is uncertain.

Each client $k$ reports $\mathrm{conf}_{k,c}$ to estimate its reliability for class $c$:
\begin{equation}
\mathrm{conf}_{k,c}
=
w_1\,\mathrm{conf}^{\mathrm{data}}_{k,c}
+
w_2\,\mathrm{conf}^{\mathrm{gen}}_{k,c}
+
w_3\,\mathrm{conf}^{\mathrm{val}}_{k,c},
\label{eq:conf_def}
\end{equation}
where $w_1+w_2+w_3=1$.
We set $\mathrm{conf}^{\mathrm{data}}_{k,c}\propto n_{k,c}^{\mathrm{eff}}$ (effective sample availability),
$\mathrm{conf}^{\mathrm{val}}_{k,c}=\exp(-\beta u_{k,c})$ using class-wise predictive uncertainty $u_{k,c}$ on a small validation split,
and $\mathrm{conf}^{\mathrm{gen}}_{k,c}$ as the average discriminator score on generated samples when tail augmentation is enabled (otherwise set to $0$).When the generator is disabled, we set $w_2=0$ and renormalize $(w_1,w_3)$ accordingly.

The server updates each global class prototype by confidence-weighted averaging:
\begin{equation}
\mathbf{p}^{t+1}_{g,c}
=
\frac{\sum_{k\in S_t}\mathrm{conf}^{t}_{k,c}\,\mathbf{p}^{t+1}_{k,c}}
{\sum_{k\in S_t}\mathrm{conf}^{t}_{k,c}+\epsilon}.
\label{eq:proto_conf_agg}
\end{equation}
Compared with naive averaging, Eq.~(\ref{eq:proto_conf_agg}) shrinks the effective squared weights of unreliable client--class pairs, thereby suppressing the variance-injection effect in Eq.~(\ref{eq:bias_rec}) and weakening cross-round error accumulation.

In addition to prototypes, we aggregate encoder parameters using an overall client confidence:
\begin{equation}
\mathrm{Conf}_k = \frac{1}{C}\sum_{c=1}^{C}\mathrm{conf}_{k,c},
\label{eq:client_conf}
\end{equation}
\begin{equation}
\theta^{t+1}
=
\frac{\sum_{k\in S_t}\mathrm{Conf}^{t}_{k}\,\theta^{t+1}_{k}}
{\sum_{k\in S_t}\mathrm{Conf}^{t}_{k}+\epsilon}.
\label{eq:enc_conf_agg}
\end{equation}
This aggregation down-weights noisy clients and improves the stability of global optimization under heterogeneous long-tailed training. For stability, we apply per-class clipping and renormalization to $\mathrm{conf}_{k,c}$.

\begin{algorithm}[t]
\caption{Confidence-Aware Federated Contrastive Learning (CAFedCL)}
\label{alg:overview_en}
\begin{algorithmic}[1]
\REQUIRE Clients $K$, classes $C$, rounds $T$; local epochs $E$; loss weights $\lambda_{\mathrm{align}},\lambda_{\mathrm{geo}}$; confidence weights $(w_1,w_2,w_3)$; small constant $\epsilon$.
\STATE Initialize global encoder $\theta^{0}$ and global prototypes $\{\mathbf{p}_{g,c}^{0}\}_{c=1}^{C}$.
\FOR{$t=0,1,\dots,T-1$}
  \STATE Sample participating clients $S_t$; broadcast $\theta^{t}$ and $\{\mathbf{p}_{g,c}^{t}\}_{c=1}^{C}$ to $k\in S_t$.
  \FOR{each client $k\in S_t$ \textbf{in parallel}}
    \STATE \textit{(Optional)} Tail augmentation; obtain augmented data $\mathcal{D}_k^{\mathrm{aug}}$ (Sec.~\ref{sec:stabilizers}).
    \STATE Update local encoder $\theta_{k}^{t+1}$ by minimizing Eq.~(\ref{eq:client_obj}) for $E$ epochs on $\mathcal{D}_k^{\mathrm{aug}}$.
    \STATE Compute local prototypes $\{\mathbf{p}_{k,c}^{t+1}\}_{c=1}^{C}$ via Eq.~(\ref{eq:local_proto}).
    \STATE Compute class-wise confidences $\{\mathrm{conf}^{t}_{k,c}\}_{c=1}^{C}$ via Eq.~(\ref{eq:conf_def}) and set $\mathrm{Conf}_k^{t}=\frac{1}{C}\sum_{c=1}^{C}\mathrm{conf}^{t}_{k,c}$.
    \STATE Upload $\theta_{k}^{t+1}$, $\{\mathbf{p}_{k,c}^{t+1}\}_{c=1}^{C}$, and $\{\mathrm{conf}^{t}_{k,c}\}_{c=1}^{C}$ to the server.
  \ENDFOR
  \STATE Update global prototypes $\{\mathbf{p}_{g,c}^{t+1}\}_{c=1}^{C}$ via Eq.~(\ref{eq:proto_conf_agg}).
  \STATE Update global encoder $\theta^{t+1}$ via Eq.~(\ref{eq:enc_conf_agg}).
\ENDFOR
\end{algorithmic}
\end{algorithm}

\subsection{Additional Stabilizers: Geometry Regularization and Tail Augmentation}
\label{sec:stabilizers}

Confidence-weighted aggregation is the primary mechanism that suppresses unreliable prototype injection across rounds.
Beyond aggregation, CAFedCL further improves minority representations with two lightweight stabilizers.
(1) a prototype alignment term that reduces coordinate mismatch between local and global prototypes;
(2) a geometry regularizer that maintains inter-class separation and prevents class collapse.
When tail supervision is extremely scarce, we additionally enable an \emph{optional} tail augmentation module to increase effective sample size.

On each client $k$, we optimize the encoder with the combined objective:
\begin{equation}
\mathcal{L}_{E,k}
=
\mathcal{L}_{\mathrm{I2P},k}
+
\lambda_{\mathrm{align}}\,\mathcal{L}_{\mathrm{align},k}
+
\lambda_{\mathrm{geo}}\,\mathcal{L}_{\mathrm{geo},k},
\label{eq:client_obj}
\end{equation}
where $\mathcal{L}_{\mathrm{I2P},k}$ is the instance-to-prototype loss in Eq.~(\ref{eq:i2p_loss}),
$\mathcal{L}_{\mathrm{align},k}$ softly aligns local prototypes to the global prototype space,
and $\mathcal{L}_{\mathrm{geo},k}$ enforces geometry constraints to prevent class collapse.
In practice, we use a moderate (or annealed) $\lambda_{\mathrm{align}}$ to avoid over-committing to early biased anchors.

The alignment term provides a gentle pull that keeps local prototypes in the same coordinate system as the broadcast global prototypes:
\begin{equation}
\mathcal{L}_{\mathrm{align},k}
=
\sum_{c=1}^{C} \mathbb{I}[n_{k,c}>0]\,
\bigl\|\mathbf{p}^{t+1}_{k,c}-\mathbf{p}^{t}_{g,c}\bigr\|_2^2 .
\label{eq:align_loss}
\end{equation}

The geometry regularizer explicitly preserves inter-class separation among prototypes, which is crucial when majority classes dominate contrastive gradients.
We instantiate $\mathcal{L}_{\mathrm{geo},k}$ with a margin-based penalty over the global prototypes:
\begin{equation}
\mathcal{L}_{\mathrm{geo},k}
=
\sum_{c\neq c'} \max\!\Bigl(0,\; m - \bigl\|\mathbf{p}_{g,c}-\mathbf{p}_{g,c'}\bigr\|_2 \Bigr).
\label{eq:geo_loss}
\end{equation}
Other geometry priors (e.g., orthogonality or simplex-like arrangements) can be used as drop-in replacements.

When minority supervision is extremely scarce, we optionally augment tail classes with a lightweight conditional generator to provide additional labeled samples and increase $n_{k,c}^{\mathrm{eff}}$.
This module is decoupled from CAFedCL and can be replaced by simpler strategies such as re-sampling or MixUp.
For completeness, we summarize the adversarial objective in the appendix and only use generated samples as additional supervised data during encoder training.

Overall, tail augmentation improves tail prototype reliability by increasing effective supervision,
while geometry regularization maintains a discriminative prototype space against majority-induced squeezing.
Together with confidence-weighted aggregation, they provide stable anchors for the next-round instance-to-prototype contrast in Eq.~(\ref{eq:i2p_loss}).

\section{Experiment}
\subsection{Experimental Setup}

\begin{figure}[t]
\centering
\includegraphics[width=\linewidth]{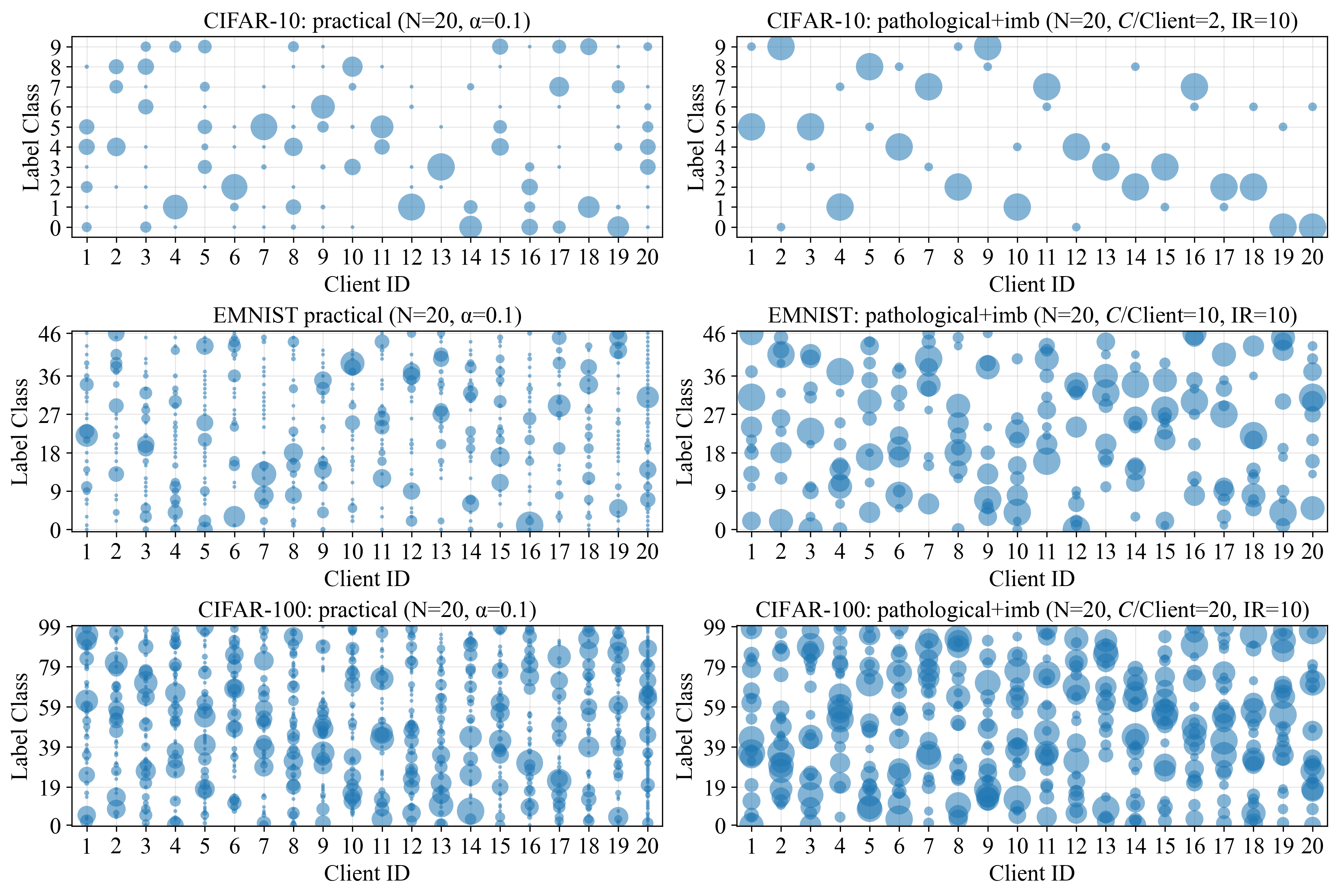}
\caption{The data distributions of CIFAR-10, EMNIST, and CIFAR-100 in the default settings.}
\label{fig: distribution of datasets}
\end{figure}

\textbf{Datasets and baselines} We employ three standard benchmarks for experiments: CIFAR-10, CIFAR-100~\citep{krizhevsky2009learning}, and EMNIST~\citep{cohen2017emnist}, covering various levels of data heterogeneity and participation rates. For non-i.i.d.\ cases, as shown in Fig.~\ref{fig: distribution of datasets}, we consider two regimes: (1) \emph{practical} setting, where label skew is simulated by sampling client-wise label proportions from a Dirichlet distribution with a symmetric parameter $\alpha$; and (2) \emph{pathological} setting, which enforces a hard limit on the number of label categories each client can receive, leading to more extreme label distribution heterogeneity. Different from prior settings, we further introduce class imbalance on top of the pathological split by controlling the imbalance ratio $\mathrm{IR}$ (the ratio between the largest and smallest class sizes). For evaluation, we use the complete test set for each dataset and report top-1 accuracy averaged over five runs. We compare our method with representative federated contrastive learning baselines, including FedAvg~\citep{mcmahan2017communication}, FedProx~\citep{li2020federated}, MOON~\citep{li2021model}, FedProto~\citep{tan2022fedproto}, FedRCL~\citep{seo2024relaxed}, FedProc~\citep{mu2023fedproc}, FedLC~\citep{zhang2022federated}, MP-FedCL~\citep{qiao2023mp}, and FedTGP~\citep{zhang2024fedtgp}.

\textbf{Implementation details} We adopt ResNet-18 as the backbone, where batch normalization is replaced by group normalization. We train the model from scratch using SGD with a learning rate of 0.1, an exponential decay parameter of 0.998, a weight decay of 0.001, and no momentum. The number of local training epochs is set to 5, and the batch size is adjusted to ensure a total of 10 local iterations per epoch throughout all experiments. We implement all methods in PyTorch and run experiments on an Intel i9-13900KF CPU with an NVIDIA GeForce RTX 4090 GPU.


\subsection{Performance}

\begin{table*}[t]
\centering
\caption{Test accuracy (\%) on three datasets under different heterogeneous settings.
Results are reported as accuracy and standard deviation over 5 runs.}
\label{tab:three_datasets_alpha_grouped}
\scriptsize
\setlength{\tabcolsep}{5pt}
\resizebox{\textwidth}{!}{%
\begin{tabular}{l|c|c|c|c|c|c|c|c|c|c|c|c}
\toprule

& \multicolumn{6}{c|}{Practical heterogeneous ($\alpha = 0.1,\, K = 20$)}
& \multicolumn{6}{c}{Pathological heterogeneous ($K = 20,\, IR = 10$)} \\
\cmidrule(lr){2-7}\cmidrule(lr){8-13} 
Method & \multicolumn{2}{c}{CIFAR-10} & \multicolumn{2}{c}{CIFAR-100} & \multicolumn{2}{c|}{EMNIST}
& \multicolumn{2}{c}{CIFAR-10} & \multicolumn{2}{c}{CIFAR-100} & \multicolumn{2}{c}{EMNIST} \\
\cmidrule(lr){2-3}\cmidrule(lr){4-5}\cmidrule(lr){6-7}
\cmidrule(lr){8-9}\cmidrule(lr){10-11}\cmidrule(lr){12-13}
& Acc.\ensuremath{\uparrow} & Std.\ensuremath{\downarrow}
& Acc.\ensuremath{\uparrow} & Std.\ensuremath{\downarrow}
& Acc.\ensuremath{\uparrow} & Std.\ensuremath{\downarrow}
& Acc.\ensuremath{\uparrow} & Std.\ensuremath{\downarrow}
& Acc.\ensuremath{\uparrow} & Std.\ensuremath{\downarrow}
& Acc.\ensuremath{\uparrow} & Std.\ensuremath{\downarrow} \\
\midrule
FedAvg
& 60.32 & 12.03 & 29.74 & 4.08 & 82.95 & 11.59
& 48.73 & 19.15 & 24.47 & 7.58 & 72.38 & 14.96 \\
FedProx
& 61.82 & 10.07 & 31.42 & 4.25 & 84.27 & 8.13
& 60.29 & 9.11  & 30.60 & 5.49 & 72.45 & 7.21 \\
MOON
& 82.06 & 11.19 & 46.23 & 5.15 & 92.77 & 6.53
& 84.94 & 13.44 & 48.70 & 6.95 & 93.24 & 5.40 \\
FedProto
& 88.33 & 8.13  & 52.81 & 6.58 & 95.50 & 1.69
& 89.45 & 7.78  & 52.14 & 5.90 & 94.88 & 2.63 \\
FedRCL
& 87.26 & 8.79  & 49.99 & 8.29 & 93.57 & 3.33
& 86.09 & 9.30  & 47.73 & 8.21 & 91.25 & 5.01 \\
FedProc
& 44.53 & 6.17  & 18.98 & 5.62 & 65.66 & 3.57
& 38.74 & 6.86  & 17.04 & 3.19 & 63.20 & 3.71 \\
FedLC
& 78.48 & 7.10  & 44.75 & 4.70 & 90.45 & 7.08
& 80.11 & 10.51 & 46.34 & 6.61 & 90.85 & 7.80 \\
MP\mbox{-}FedCL
& 86.94 & 8.46  & 49.70 & 6.26 & 94.30 & 3.28
& 87.37 & 7.55  & 47.27 & 5.98 & 90.05 & 4.33 \\
FedTGP
& 89.24 & 7.03  & 51.30 & 8.59 & 94.47 & 4.54
& 89.57 & 8.46  & 51.65 & 6.97 & 94.59 & 4.03 \\
\midrule
\textbf{CAFedCL}
& \textbf{91.15} & \textbf{4.87} & \textbf{54.01} & \textbf{3.65} & \textbf{96.84} & \textbf{1.54}
& \textbf{90.36} & \textbf{5.08} & \textbf{54.57} & \textbf{2.82} & \textbf{97.00} & \textbf{2.25} \\
\bottomrule
\end{tabular}
}%
\end{table*}

We compare the proposed method, CAFedCL, with recent federated contrastive learning baselines and representative FL methods on the CIFAR and EMNIST datasets. Table~\ref{tab:three_datasets_alpha_grouped} shows that CAFedCL consistently outperforms state-of-the-art baselines across all datasets and heterogeneous settings. Classical optimization-based methods such as FedAvg and FedProx experience pronounced performance drops under heterogeneity and exhibit large client-wise dispersion. Here, std denotes the standard deviation of client test accuracies, where a smaller value indicates smaller accuracy gaps among clients and thus better fairness. Recent contrastive baselines (e.g., MOON and FedRCL) yield noticeable improvements, but their gains are less stable on fine-grained tasks such as CIFAR-100. Prototype-based FedProto partially improves representation alignment, yet it remains vulnerable to the prototype bias loop under extreme imbalance. In contrast, CAFedCL achieves both the highest global accuracy and the lowest client-wise Std, indicating that the proposed confidence-aware mechanism effectively downweights unreliable updates and avoids sacrificing hard-to-learn clients. Overall, CAFedCL delivers stronger generalization while markedly improving client fairness, making it more suitable for heterogeneous and imbalanced scenarios.

\subsection{Robustness}

\newcommand{\up}{\ensuremath{\uparrow}}
\newcommand{\down}{\ensuremath{\downarrow}}

\begin{table*}[t]
\centering
\setlength{\tabcolsep}{6pt} 
\caption{Performance (\%) under heterogeneous settings. Results are reported as accuracy and standard deviation over 5 runs.}
\label{tab:hetero}
\resizebox{\textwidth}{!}{%
\begin{tabular}{l|cc|cc|cc|cc|cc|cc|cc|cc}
\toprule
\multirow{3}{*}{Method} 
& \multicolumn{4}{c}{Dirichlet parameter}
& \multicolumn{4}{|c}{Imbalance ratio}
& \multicolumn{4}{|c}{Class per client}
& \multicolumn{4}{|c}{Client amount} \\
\cmidrule(lr){2-5}\cmidrule(lr){6-9}\cmidrule(lr){10-13}\cmidrule(lr){14-17}
& \multicolumn{2}{c}{$\alpha = 0.05$} & \multicolumn{2}{c}{$\alpha = 1.0$}
& \multicolumn{2}{|c}{$IR = 50$} & \multicolumn{2}{c}{$IR = 100$}
& \multicolumn{2}{|c}{$C/\text{Client} = 10$} & \multicolumn{2}{c}{$C/\text{Client} = 50$}
& \multicolumn{2}{|c}{50 Clients} & \multicolumn{2}{c}{100 Clients} \\
\cmidrule(lr){2-3}\cmidrule(lr){4-5}
\cmidrule(lr){6-7}\cmidrule(lr){8-9}
\cmidrule(lr){10-11}\cmidrule(lr){12-13}
\cmidrule(lr){14-15}\cmidrule(lr){16-17}
& Acc.\up & Std.\down & Acc.\up & Std.\down
& Acc.\up & Std.\down & Acc.\up & Std.\down
& Acc.\up & Std.\down & Acc.\up & Std.\down
& Acc.\up & Std.\down & Acc.\up & Std.\down \\
\midrule

FedAvg
& 52.54 & 12.24 & 86.67 & 6.19 
& 43.28 & 22.81 & 35.29 & 17.50 
& 20.87 & 7.12 & 29.66 & 2.59 
& 56.73 & 12.88 & 54.14 & 13.65 \\
FedProx
& 55.48 & 14.05 & 87.93 & 6.44 
& 48.30 & 11.32 & 37.33 & 13.72 
& 28.56 & 6.28 & 32.70 & 2.85 
& 58.50 & 10.12 & 55.94 & 10.47 \\
MOON
& 65.79 & 10.40 & 90.95 & 5.52 
& 54.23 & 15.73 & 42.61 & 18.25 
& 56.33 & 4.95 & 25.49 & 5.22 
& 79.51 & 11.58 & 78.19 & 10.66 \\
FedProto
& 66.86 & 8.35 & 91.27 & 5.24 
& 55.48 & 9.56 & 44.46 & 10.29 
& 62.12 & 4.66 & 35.06 & 4.41 
& 83.07 & 9.92 & 80.44 & 10.25 \\
FedRCL
& 68.30 & 9.83 & 92.56 & 4.77 
& 53.93 & 10.36 & 43.08 & 10.78 
& 60.29 & 3.98 & 34.77 & 2.54 
& 84.08 & 8.73 & 82.75 & 9.11 \\

FedProc
& 35.68 & 9.11 & 84.33 & 6.80 
& 30.65 & 8.41 & 25.49 & 11.97 
& 29.81 & 4.02 & 15.88 & 4.45 
& 42.24 & 7.02 & 41.52 & 7.31 \\
FedLC
& 67.33 & 7.26 & 91.90 & 5.03 
& 53.66 & 13.08 & 42.41 & 9.28 
& 55.09 & 5.17 & 34.70 & 3.32 
& 75.32 & 7.60 & 73.09 & 8.51 \\
MP-FedCL
& 66.18 & 6.79 & 91.21 & 4.52 
& 57.21 & 9.37 & 45.15 & 10.63 
& 62.94 & 3.19 & 36.59 & 3.21 
& 82.25 & 10.20 & 79.62 & 10.55 \\
FedTGP
& 67.14 & 7.01 & 92.58 & 4.33 
& 58.67 & 7.55 & 45.74 & 7.82 
& 64.16 & 4.66 & 36.41 & 2.76 
& 83.77 & 8.24 & 81.95 & 8.90 \\
\midrule
\textbf{CAFedCL}
& \textbf{70.22} & \textbf{5.17} & \textbf{93.96} & \textbf{4.07}
& \textbf{60.14} & \textbf{5.58} & \textbf{46.53} & \textbf{6.47}
& \textbf{65.85} & \textbf{2.56} & \textbf{39.22} & \textbf{2.48}
& \textbf{85.16} & \textbf{5.92} & \textbf{83.60} & \textbf{6.39} \\
\bottomrule
\end{tabular}%
}
\end{table*}

To further evaluate robustness, we vary key heterogeneity and scale factors, including the Dirichlet parameter $\alpha$, imbalance ratio $\mathrm{IR}$, the number of classes per client ($C$/Client), and the total client amount. In this table, \textbf{Std} denotes the standard deviation of client accuracies; thus, a smaller Std indicates a smaller performance gap among clients and better fairness. Across all tested conditions, CAFedCL achieves the best accuracy and the lowest client-wise Std, demonstrating consistently strong effectiveness and fairness.

When label heterogeneity becomes more severe (e.g., $\alpha=0.05$), CAFedCL remains noticeably more stable than prior methods, reaching 70.22 accuracy with Std 5.17, while many baselines exhibit much larger dispersion. As the class imbalance intensifies, CAFedCL degrades gracefully and continues to be the most reliable, obtaining 60.14/46.53 accuracy under $\mathrm{IR}=50/100$ with relatively low Std (5.58/6.47), whereas classical methods suffer from both sharp accuracy drops and inflated variance. Under varying $C$/Client constraints, CAFedCL consistently yields the smallest client-wise gap (Std 2.56 and 2.48), indicating strong fairness even under extreme class-sparsity regimes. Finally, when scaling the system to more clients (50 $\rightarrow$ 100), CAFedCL maintains competitive performance (85.16 $\rightarrow$ 83.60) with limited increase in Std (5.92 $\rightarrow$ 6.39), suggesting good scalability. Overall, CAFedCL provides robust improvements without sacrificing hard-to-learn clients, leading to consistently higher accuracy \emph{and} better client fairness under diverse heterogeneous and imbalanced settings.

\subsection{Ablation Study}
\begin{figure}[t]
\centering
\includegraphics[width=\columnwidth]{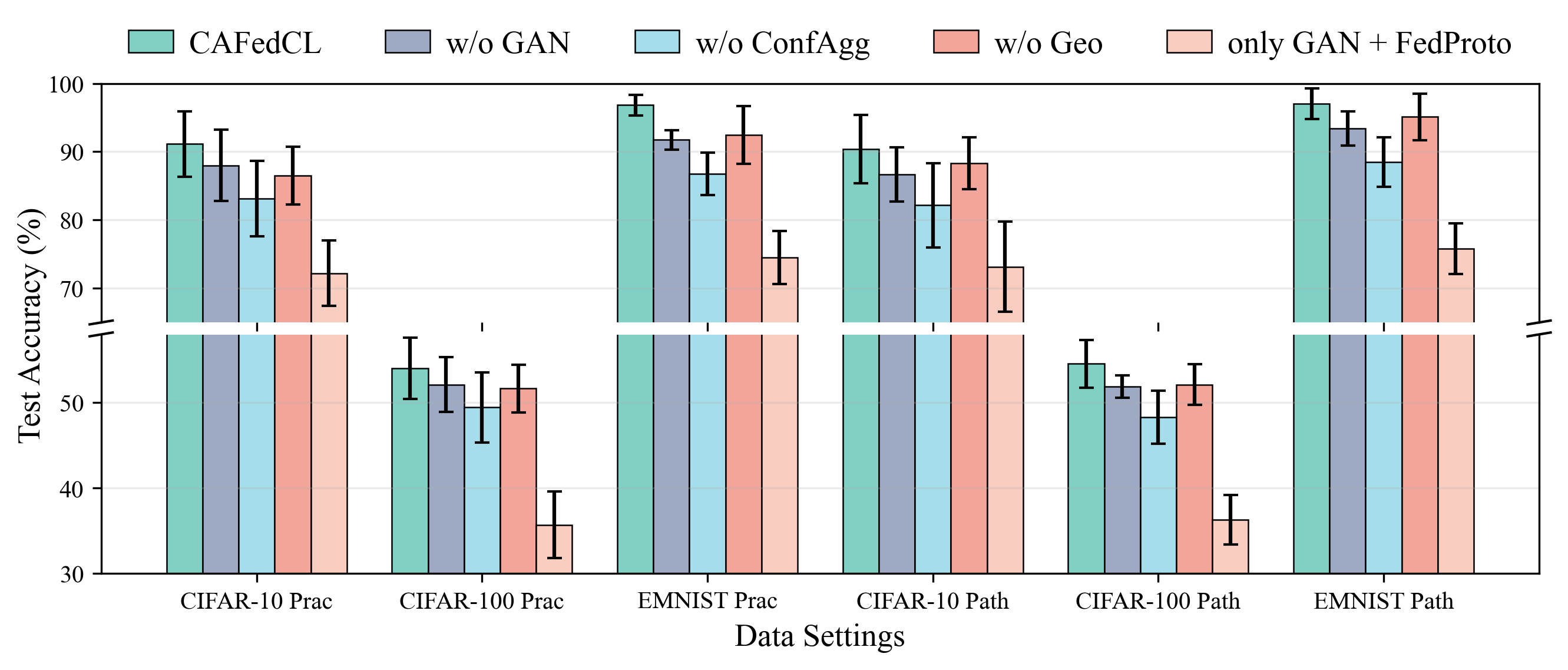}
\caption{The accuracy of CAFedCL and its ablated variants under the default settings.}
\label{fig: ablation}
\end{figure}

We report an ablation study in Fig.~\ref{fig: ablation} to isolate the contribution of each component in CAFedCL across six settings (three datasets under practical and pathological heterogeneity). We consider four variants: (i) w/o GAN removes the minority-class generative augmentation; (ii) w/o ConfAgg replaces our confidence-aware aggregation with a naive aggregation scheme; (iii) w/o Geo drops the geometric consistency regularization; and (iv) only GAN + FedProto keeps only GAN-based augmentation on top of FedProto, discarding the aggregation and regularization designs of CAFedCL. Overall, CAFedCL consistently achieves the best accuracy, while all variants suffer performance degradation, indicating that the improvements stem from the synergy of these components. Among them, w/o ConfAgg yields the largest and most consistent drop (e.g., CIFAR-10: 91.15$\rightarrow$83.13 in practical and 90.36$\rightarrow$82.15 in pathological; EMNIST: 96.84$\rightarrow$86.77 and 97.00$\rightarrow$88.48), highlighting that confidence-aware aggregation is crucial for suppressing unreliable clients and preventing biased prototypes from accumulating across rounds. Removing GAN augmentation also leads to clear degradation, especially under more challenging heterogeneity, confirming its role in improving minority coverage and stabilizing prototype estimates. w/o Geo causes a moderate yet non-negligible drop, suggesting that geometric regularization helps maintain a consistent class structure under client drift. Finally, only GAN + FedProto performs substantially worse than CAFedCL across all settings, demonstrating that simply injecting synthetic samples into a prototype baseline is insufficient without ConfAgg and the proposed representation-level constraints.

\subsection{Hyperparameter Sensitivity Analysis}
\begin{figure}[t]
\centering
\includegraphics[width=\columnwidth]{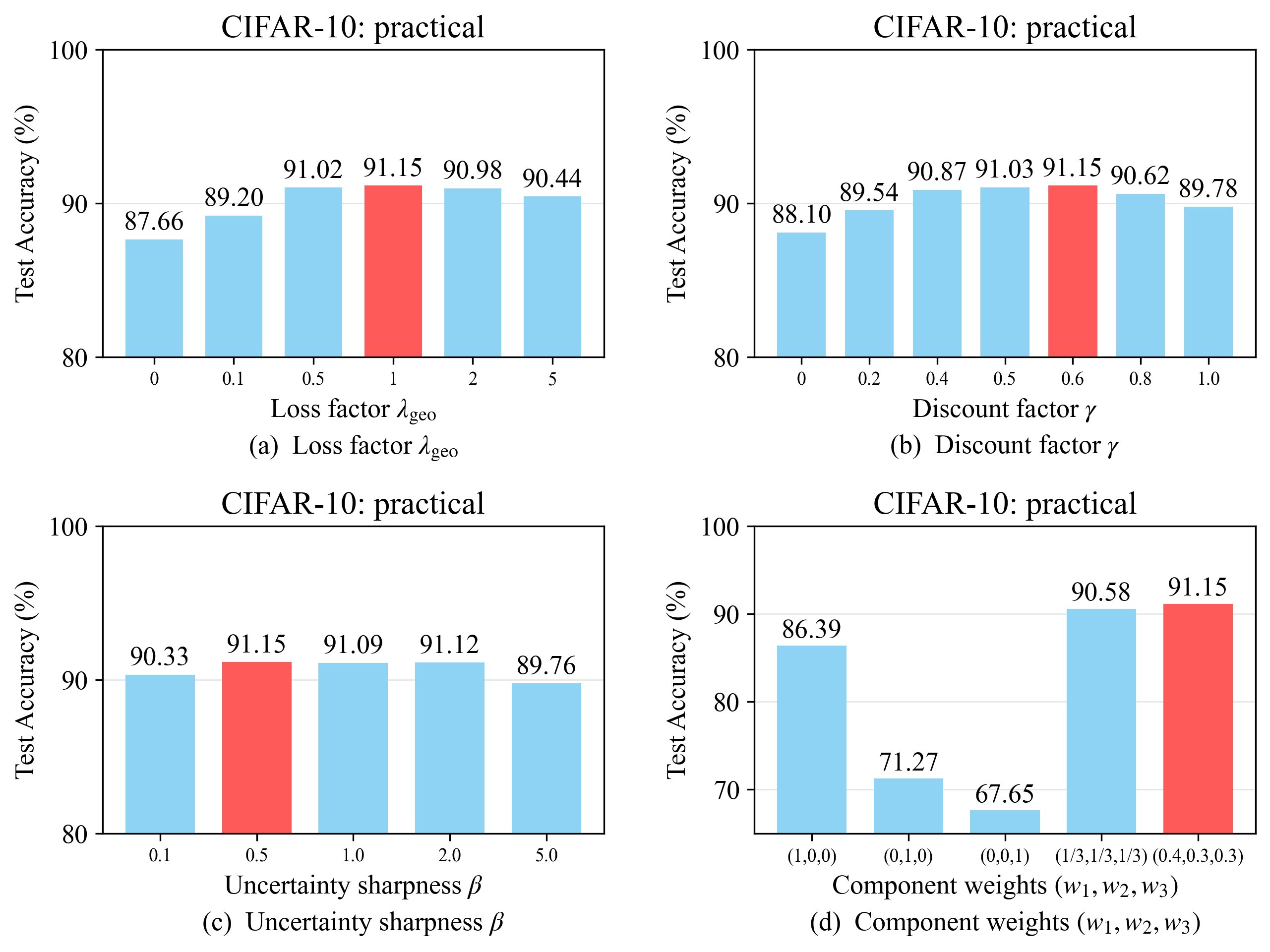}
\caption{The accuracy of CAFedCL with different hyperparameters.}
\label{fig: hyperparam}
\end{figure}

Fig.~\ref{fig: hyperparam} reports the hyperparameter sensitivity of \textbf{CAFedCL} on CIFAR-10 under the practical heterogeneous setting. Overall, CAFedCL is stable across a reasonably wide range and consistently favors \emph{moderate} parameter choices. Increasing the geometric loss weight $\lambda_{\mathrm{geo}}$ improves accuracy from 87.66 ($\lambda_{\mathrm{geo}}{=}0$) to a peak of 91.15 at $\lambda_{\mathrm{geo}}{=}1$, while larger values slightly degrade performance, indicating that overly strong geometric regularization can over-constrain optimization. The synthetic discount factor $\gamma$ exhibits a similar ``moderate-is-best'' trend, reaching the best accuracy at $\gamma{=}0.6$ (91.15) and decreasing as $\gamma$ approaches 1.0, which supports discounting generated samples to avoid over-trusting imperfect generations. For uncertainty sharpness $\beta$, performance peaks at $\beta{=}0.5$ (91.15) and remains robust for $\beta\in[0.5,2]$, whereas too small or too large $\beta$ weakens discrimination or makes aggregation overly selective. Finally, varying the confidence component weights $(w_1,w_2,w_3)$ shows that relying on a single component is suboptimal (e.g., $(0,0,1)$: 67.65), while combining complementary signals yields strong and stable performance, with our default mixture $(0.4,0.3,0.3)$ achieving the best result.


\section{Conclusion}
\label{sec:conclusion}

This study addresses the critical challenges of federated contrastive learning under extreme class imbalance and client heterogeneity. Although prototype-based communication offers communication efficiency, it is inherently vulnerable to instability. We formally identify and characterize a prototype bias loop: noisy and biased client prototypes aggregate into inaccurate global anchors, which are subsequently reinforced by prototype-guided contrastive objectives, inducing cumulative representation degradation across communication rounds. To disrupt this vicious cycle, we propose CAFedCL (Confidence-Aware Federated Contrastive Learning), featuring three synergistic components: (1) class-wise reliability weighting for dynamically calibrating both prototype and model aggregation; (2) global geometry regularization to preserve structural integrity of the prototype space; and (3) tail-focused augmentation to explicitly strengthen minority-class representations. This integrated design enhances system robustness and inter-client fairness without increasing communication overhead.

Several directions are worth exploring in future work. First, developing more lightweight and computationally efficient class-wise reliability estimation strategies could further improve the scalability of CAFedCL in large-scale federated systems. Second, integrating confidence-aware aggregation with privacy-preserving techniques, such as differential privacy or secure aggregation, is crucial for practical deployment. Finally, extending the proposed framework to more complex federated settings, including heterogeneous model architectures or asynchronous communication, remains an interesting direction for future investigation, especially to understand how model mismatch and time-varying client participation affect confidence calibration and the stability of prototype alignment in open-world deployments.

\bibliography{main}

@ARTICLE{wang2025lightweight,
  author={Wang, Rui and Huang, Weiguo and Huang, Wenke},
  journal={IEEE Internet of Things Journal}, 
  title={Lightweight Federated Domain Generalization With Global–Local Contrastive Learning for Machine Fault Diagnosis}, 
  year={2025},
  volume={12},
  number={19},
  pages={40750--40763},
  doi={10.1109/JIOT.2025.3590721}
}

@article{qiao2023mp,
  title={Mp-fedcl: Multiprototype federated contrastive learning for edge intelligence},
  author={Qiao, Yu and Munir, Md Shirajum and Adhikary, Apurba and Le, Huy Q and Raha, Avi Deb and Zhang, Chaoning and Hong, Choong Seon},
  journal={IEEE Internet of Things journal},
  volume={11},
  number={5},
  pages={8604--8623},
  year={2023},
  publisher={IEEE}
}

@article{tan2022federated,
  title={Federated learning from pre-trained models: A contrastive learning approach},
  author={Tan, Yue and Long, Guodong and Ma, Jie and Liu, Lu and Zhou, Tianyi and Jiang, Jing},
  journal={Advances in neural information processing systems},
  volume={35},
  pages={19332--19344},
  year={2022}
}

@inproceedings{wu2023fediic,
  title={Fediic: Towards robust federated learning for class-imbalanced medical image classification},
  author={Wu, Nannan and Yu, Li and Yang, Xin and Cheng, Kwang-Ting and Yan, Zengqiang},
  booktitle={International Conference on Medical Image Computing and Computer-Assisted Intervention},
  pages={692--702},
  year={2023},
  organization={Springer}
}

@article{huang2025fedcosr,
  title={FedCoSR: Personalized Federated Learning With Contrastive Shareable Representations for Label Heterogeneity in Non-IID Data},
  author={Huang, Chenghao and Chen, Xiaolu and Zhang, Yanru and Wang, Hao},
  journal={IEEE Transactions on Cybernetics},
  year={2025},
  publisher={IEEE}
}

@inproceedings{zhang2024fedtgp,
  title={Fedtgp: Trainable global prototypes with adaptive-margin-enhanced contrastive learning for data and model heterogeneity in federated learning},
  author={Zhang, Jianqing and Liu, Yang and Hua, Yang and Cao, Jian},
  booktitle={Proceedings of the AAAI conference on artificial intelligence},
  volume={38},
  number={15},
  pages={16768--16776},
  year={2024}
}

@inproceedings{mcmahan2017communication,
  title={Communication-efficient learning of deep networks from decentralized data},
  author={McMahan, Brendan and Moore, Eider and Ramage, Daniel and Hampson, Seth and y Arcas, Blaise Aguera},
  booktitle={Artificial intelligence and statistics},
  pages={1273--1282},
  year={2017},
  organization={PMLR}
}

@article{kairouz2021advances,
  title={Advances and open problems in federated learning},
  author={Kairouz, Peter and McMahan, H Brendan and Avent, Brendan and Bellet, Aur{\'e}lien and Bennis, Mehdi and Bhagoji, Arjun Nitin and Bonawitz, Kallista and Charles, Zachary and Cormode, Graham and Cummings, Rachel and others},
  journal={Foundations and trends{\textregistered} in machine learning},
  volume={14},
  number={1--2},
  pages={1--210},
  year={2021},
  publisher={Now Publishers, Inc.}
}

@article{imteaj2021survey,
  title={A survey on federated learning for resource-constrained IoT devices},
  author={Imteaj, Ahmed and Thakker, Urmish and Wang, Shiqiang and Li, Jian and Amini, M Hadi},
  journal={IEEE Internet of Things Journal},
  volume={9},
  number={1},
  pages={1--24},
  year={2021},
  publisher={IEEE}
}

@article{tan2022towards,
  title={Towards personalized federated learning},
  author={Tan, Alysa Ziying and Yu, Han and Cui, Lizhen and Yang, Qiang},
  journal={IEEE transactions on neural networks and learning systems},
  volume={34},
  number={12},
  pages={9587--9603},
  year={2022},
  publisher={IEEE}
}

@article{mu2023fedproc,
  title={Fedproc: Prototypical contrastive federated learning on non-iid data},
  author={Mu, Xutong and Shen, Yulong and Cheng, Ke and Geng, Xueli and Fu, Jiaxuan and Zhang, Tao and Zhang, Zhiwei},
  journal={Future Generation Computer Systems},
  volume={143},
  pages={93--104},
  year={2023},
  publisher={Elsevier}
}

@inproceedings{tan2022fedproto,
  title={Fedproto: Federated prototype learning across heterogeneous clients},
  author={Tan, Yue and Long, Guodong and Liu, Lu and Zhou, Tianyi and Lu, Qinghua and Jiang, Jing and Zhang, Chengqi},
  booktitle={Proceedings of the AAAI conference on artificial intelligence},
  volume={36},
  number={8},
  pages={8432--8440},
  year={2022}
}

@inproceedings{li2021model,
  title={Model-contrastive federated learning},
  author={Li, Qinbin and He, Bingsheng and Song, Dawn},
  booktitle={Proceedings of the IEEE/CVF conference on computer vision and pattern recognition},
  pages={10713--10722},
  year={2021}
}

@article{li2020federated,
  title={Federated optimization in heterogeneous networks},
  author={Li, Tian and Sahu, Anit Kumar and Zaheer, Manzil and Sanjabi, Maziar and Talwalkar, Ameet and Smith, Virginia},
  journal={Proceedings of Machine learning and systems},
  volume={2},
  pages={429--450},
  year={2020}
}

@article{luo2021no,
  title={No fear of heterogeneity: Classifier calibration for federated learning with non-iid data},
  author={Luo, Mi and Chen, Fei and Hu, Dapeng and Zhang, Yifan and Liang, Jian and Feng, Jiashi},
  journal={Advances in Neural Information Processing Systems},
  volume={34},
  pages={5972--5984},
  year={2021}
}

@inproceedings{wang2021addressing,
  title={Addressing class imbalance in federated learning},
  author={Wang, Lixu and Xu, Shichao and Wang, Xiao and Zhu, Qi},
  booktitle={Proceedings of the AAAI conference on artificial intelligence},
  volume={35},
  number={11},
  pages={10165--10173},
  year={2021}
}

@inproceedings{diao2024exploiting,
  title={Exploiting label skews in federated learning with model concatenation},
  author={Diao, Yiqun and Li, Qinbin and He, Bingsheng},
  booktitle={Proceedings of the AAAI Conference on Artificial Intelligence},
  volume={38},
  number={10},
  pages={11784--11792},
  year={2024}
}

@article{wang2020tackling,
  title={Tackling the objective inconsistency problem in heterogeneous federated optimization},
  author={Wang, Jianyu and Liu, Qinghua and Liang, Hao and Joshi, Gauri and Poor, H Vincent},
  journal={Advances in neural information processing systems},
  volume={33},
  pages={7611--7623},
  year={2020}
}

@inproceedings{karimireddy2020scaffold,
  title={Scaffold: Stochastic controlled averaging for federated learning},
  author={Karimireddy, Sai Praneeth and Kale, Satyen and Mohri, Mehryar and Reddi, Sashank and Stich, Sebastian and Suresh, Ananda Theertha},
  booktitle={International conference on machine learning},
  pages={5132--5143},
  year={2020},
  organization={PMLR}
}

@inproceedings{li2021adaptive,
  title={Adaptive prototype learning and allocation for few-shot segmentation},
  author={Li, Gen and Jampani, Varun and Sevilla-Lara, Laura and Sun, Deqing and Kim, Jonghyun and Kim, Joongkyu},
  booktitle={Proceedings of the IEEE/CVF conference on computer vision and pattern recognition},
  pages={8334--8343},
  year={2021}
}

@article{li2021fedbn,
  title={Fedbn: Federated learning on non-iid features via local batch normalization},
  author={Li, Xiaoxiao and Jiang, Meirui and Zhang, Xiaofei and Kamp, Michael and Dou, Qi},
  journal={arXiv preprint arXiv:2102.07623},
  year={2021}
}

@inproceedings{dai2023tackling,
  title={Tackling data heterogeneity in federated learning with class prototypes},
  author={Dai, Yutong and Chen, Zeyuan and Li, Junnan and Heinecke, Shelby and Sun, Lichao and Xu, Ran},
  booktitle={Proceedings of the AAAI Conference on Artificial Intelligence},
  volume={37},
  number={6},
  pages={7314--7322},
  year={2023}
}

@inproceedings{seo2024relaxed,
  title={Relaxed contrastive learning for federated learning},
  author={Seo, Seonguk and Kim, Jinkyu and Kim, Geeho and Han, Bohyung},
  booktitle={Proceedings of the IEEE/CVF Conference on Computer Vision and Pattern Recognition},
  pages={12279--12288},
  year={2024}
}

@article{zhao2018federated,
  title={Federated learning with non-iid data},
  author={Zhao, Yue and Li, Meng and Lai, Liangzhen and Suda, Naveen and Civin, Damon and Chandra, Vikas},
  journal={arXiv preprint arXiv:1806.00582},
  year={2018}
}

@article{zhuang2022divergence,
  title={Divergence-aware federated self-supervised learning},
  author={Zhuang, Weiming and Wen, Yonggang and Zhang, Shuai},
  journal={arXiv preprint arXiv:2204.04385},
  year={2022}
}

@inproceedings{guo2023fedbr,
  title={Fedbr: Improving federated learning on heterogeneous data via local learning bias reduction},
  author={Guo, Yongxin and Tang, Xiaoying and Lin, Tao},
  booktitle={International conference on machine learning},
  pages={12034--12054},
  year={2023},
  organization={PMLR}
}

@inproceedings{jing2024fedsc,
author = {Jing, Shusen and Yu, Anlan and Zhang, Shuai and Zhang, Songyang},
title = {FedSC: provable federated self-supervised learning with spectral contrastive objective over non-i.i.d. data},
year = {2024},
publisher = {JMLR.org},
booktitle = {Proceedings of the 41st International Conference on Machine Learning},
articleno = {897},
numpages = {22},
location = {Vienna, Austria},
series = {ICML'24}
}

@inproceedings{zhang2021federated,
  title={Federated learning for non-iid data via unified feature learning and optimization objective alignment},
  author={Zhang, Lin and Luo, Yong and Bai, Yan and Du, Bo and Duan, Ling-Yu},
  booktitle={Proceedings of the IEEE/CVF international conference on computer vision},
  pages={4420--4428},
  year={2021}
}

@inproceedings{chen2020simple,
  title={A simple framework for contrastive learning of visual representations},
  author={Chen, Ting and Kornblith, Simon and Norouzi, Mohammad and Hinton, Geoffrey},
  booktitle={International conference on machine learning},
  pages={1597--1607},
  year={2020},
  organization={PmLR}
}

@inproceedings{he2020momentum,
  title={Momentum contrast for unsupervised visual representation learning},
  author={He, Kaiming and Fan, Haoqi and Wu, Yuxin and Xie, Saining and Girshick, Ross},
  booktitle={Proceedings of the IEEE/CVF conference on computer vision and pattern recognition},
  pages={9729--9738},
  year={2020}
}

@article{oord2018representation,
  title={Representation learning with contrastive predictive coding},
  author={Oord, Aaron van den and Li, Yazhe and Vinyals, Oriol},
  journal={arXiv preprint arXiv:1807.03748},
  year={2018}
}

@article{khan2021federated,
  title={Federated learning for internet of things: Recent advances, taxonomy, and open challenges},
  author={Khan, Latif U and Saad, Walid and Han, Zhu and Hossain, Ekram and Hong, Choong Seon},
  journal={IEEE Communications Surveys \& Tutorials},
  volume={23},
  number={3},
  pages={1759--1799},
  year={2021},
  publisher={IEEE}
}

@article{adnan2022federated,
  title={Federated learning and differential privacy for medical image analysis},
  author={Adnan, Mohammed and Kalra, Shivam and Cresswell, Jesse C and Taylor, Graham W and Tizhoosh, Hamid R},
  journal={Scientific reports},
  volume={12},
  number={1},
  pages={1953},
  year={2022},
  publisher={Nature Publishing Group UK London}
}

@inproceedings{liu2021feddg,
  title={Feddg: Federated domain generalization on medical image segmentation via episodic learning in continuous frequency space},
  author={Liu, Quande and Chen, Cheng and Qin, Jing and Dou, Qi and Heng, Pheng-Ann},
  booktitle={Proceedings of the IEEE/CVF conference on computer vision and pattern recognition},
  pages={1013--1023},
  year={2021}
}

@inproceedings{cohen2017emnist,
  title={EMNIST: Extending MNIST to handwritten letters},
  author={Cohen, Gregory and Afshar, Saeed and Tapson, Jonathan and Van Schaik, Andre},
  booktitle={2017 international joint conference on neural networks (IJCNN)},
  pages={2921--2926},
  year={2017},
  organization={IEEE}
}

@article{krizhevsky2009learning,
  title={Learning multiple layers of features from tiny images},
  author={Krizhevsky, Alex and Hinton, Geoffrey and others},
  year={2009},
  publisher={Toronto, ON, Canada}
}

@inproceedings{zhang2022federated,
  title={Federated learning with label distribution skew via logits calibration},
  author={Zhang, Jie and Li, Zhiqi and Li, Bo and Xu, Jianghe and Wu, Shuang and Ding, Shouhong and Wu, Chao},
  booktitle={International Conference on Machine Learning},
  pages={26311--26329},
  year={2022},
  organization={PMLR}
}

\end{document}